# ALCN: Meta-Learning for Contrast Normalization Applied to Robust 3D Pose Estimation

Mahdi Rad[1]
rad@icg.tugraz.at

Peter M. Roth[1]
pmroth@icg.tugraz.at

Vincent Lepetit[1,2]
lepetit@icg.tugraz.at

[1] Institute of Computer Graphics and Vision
Graz University of Technology
Graz, Austria

[2] Laboratoire Bordelais de Recherche en Informatique
Université de Bordeaux
Bordeaux, France

## Abstract

To be robust to illumination changes when detecting objects in images, the current trend is to train a Deep Network with training images captured under many different lighting conditions. Unfortunately, creating such a training set is very cumbersome, or sometimes even impossible, for some applications such as 3D pose estimation of specific objects, which is the application we focus on in this paper. We therefore propose a novel illumination normalization method that lets us learn to detect objects and estimate their 3D pose under challenging illumination conditions from very few training samples. Our key insight is that normalization parameters should adapt to the input image. In particular, we realized this via a Convolutional Neural Network trained to predict the parameters of a generalization of the Difference-of-Gaussians method. We show that our method significantly outperforms standard normalization methods and demonstrate it on two challenging 3D detection and pose estimation problems.

## 1 Introduction

Over the last years, Deep Networks [11, 12, 22] have spectacularly improved the performance of computer vision applications. So far, however, efforts have been mainly focused on tasks where large amounts of training data are available. To be robust to illumination conditions for example, one can train a Deep Network with many samples captured under various illumination conditions.

While for some general categories such as faces, cars, or pedestrians, training data can be exploited from other data, or capturing many images under different conditions is possible, it becomes very cumbersome for others. For example, as illustrated in Fig. 1, we want to estimate the 3D pose of specific objects without having to vary the illumination when capturing training images. To achieve this, we could use a contrast normalization technique such as Local Contrast Normalization [10], Difference-of-Gaussians or histogram normalization.

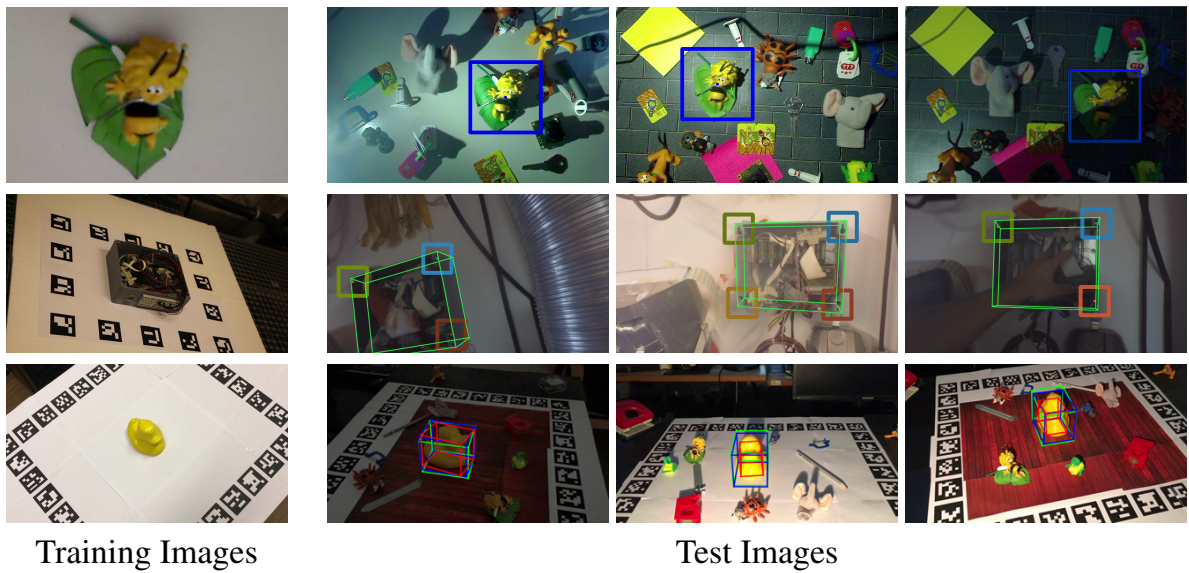

Training Images          Test Images

Figure 1: We propose a novel approach to illumination normalization, which allows us to deal with strong light changes even when only few training samples are available. We apply it to 2D detection (first row) and 3D object detection using the methods of [3] (second row) and of [19] (third row): Given training images under constant illumination, we can detect the object and predict its pose under various and drastic illumination. In the third row, green bounding boxes show the ground truth pose, blue bounding boxes represent the pose obtained with ALCN, and red bounding boxes the pose obtained without normalization.

However, our experiments show that existing methods often fail when dealing with large magnitudes of illumination changes.

Among the various existing normalization methods, Difference-of-Gaussians still performs best in our experiments, which inspired us to introduce a normalization model building on a linear combination of 2D Gaussian kernels with fixed standard deviations. But instead of using fixed parameters, we propose to adapt these parameters to the illumination conditions of the different image regions: This way, we can handle bigger illumination changes and avoid manual tuning.

However, the link between a given image and the best parameters is not straightforward. We therefore want to learn to predict these parameters from the image using a CNN. Since we do not know *a priori* the parameters to predict, we cannot train this CNN in a standard supervised manner. Our solution is to train it *jointly* in a supervised way together with another CNN to achieve object detection under illumination changes.

We call this method Adaptive Local Contrast Normalization (ALCN), as it is related to previous Local Contrast Normalization methods while being adaptive. We show that ALCN outperforms by a large margin previous methods for illumination normalization while we do not need any manual tuning. It also outperforms Deep Networks including VGG [22] and ResNet [8] trained on the same images, showing that our approach can generalize better to unseen illumination variations than a single network.

To summarize, our main contribution is an adaptive and efficient image normalization approach that makes Deep Networks more robust to illumination changes unseen during training, therefore requiring much less amounts of training data. Furthermore, we created new datasets for benchmarking of object detection and 3D pose estimation under challenging lightening conditions with distractor objects and cluttered background.

## 2 Related Work

Reliable computer vision methods require to be invariant, or at least robust, to many different visual nuisances, including pose and illumination variations. In the following, give an overview of different, sometimes complementary approaches to achieve this.

A first approach is to normalize the input image using image statistics. Several methods have been proposed, sometimes used together with Deep Networks such as SLCN and DLCN: Difference-of-Gaussians (DoG), Whitening, Subtractive and Divisive Local Contrast Normalization (SLCN and DLCN) [10], Local Response Normalization (LRN) [11], Histogram Equalization (HE), Contrast Limited Adaptive Histogram Equalization (CLAHE) [18]. However, illumination is not necessarily uniform over an image: Applying one of these methods locally over regions of the image handles local light changes better, but unfortunately they can also become unstable on poorly textured regions. Our approach overcomes this limitation with an adaptive method that effectively adjusts the normalization according to the local appearance of the image.

An alternative method is to use locally invariant features. For example, Haar wavelets [23] and the pairwise intensity comparisons used in Local Binary Patterns [17] are invariant to monotonic changes of the intensities. Features based on image gradients are invariant to constants added to the intensities. In practice, they are also often made invariant to affine changes by normalizing gradient magnitudes over the bins indexed by their orientations [14]. The SIFT descriptors are additionally normalized by an iterative process that makes them robust to saturation effects as well [15]. However, it is difficult to come up with features that are invariant to complex illumination changes on 3D objects, such as changes of light direction, cast or self shadows.

A third approach is to model illumination explicitly and estimate an intrinsic image or a self quotient image of the input image, to get rid of the illumination and isolate the reflectance of the scene as an invariant to illumination [20, 21, 25, 27]. However, it is still difficult to get an intrinsic image from one single input image that is good enough for computer vision tasks, as our experiments will show for 2D object detection.

The current trend to achieve robustness to illumination changes is to train Deep Networks with different illuminations present in the training set [22]. This, however, requires the acquisition of many training images under various conditions. As we will show, our approach performs better than single Deep Networks when illumination variations are limited in the training set, which can be the case in practice for some applications.

## 3 Adaptive Local Contrast Normalization

In this section, we first introduce our normalization model, then we discuss how we train a CNN to predict the model parameters for a given image region and how we can efficiently extend this method to a whole image.

### 3.1 Normalization Model

As our experiments in Section 4 will show, the Difference-of-Gaussians normalization method performs best among the existing normalization methods, however, it is difficult to find the standard deviations that perform well for any input image. We therefore introduce the fol-lowing formulation for our ALCN method:

$$\text{ALCN}(I;\, w) = \left( \sum_{i=1}^{N} w_i \cdot G_{\sigma_i^{\text{ALCN}}} \right) * I, \tag{1}$$

where $I$ is an input image window, $w$ a vector containing the parameters of the method, and $\text{ALCN}(I; w)$ is the normalized image; $G_\sigma$ denotes a Gaussian kernel of standard deviation $\sigma$; the $\sigma_i^{\text{ALCN}}$ are fixed standard deviations, and $*$ denotes the convolution product. In the experiments, we use ten different 2D Gaussian filters $G_{\sigma_i^{\text{ALCN}}}$, with standard deviation $\sigma_i^{\text{ALCN}} = i/2$ for $i = 1, 2, ..., 10$. This model is a generalization of the Difference-of-Gaussians model, since the normalized image is obtained by convolution of a linear combination of Gaussian kernels, and the weights of this linear combination are the parameters of the model.

## 3.2 Joint Training to Predict the Model Parameters

As discussed in the introduction and shown in Fig. 2(a), we train a Convolutional Neural Network (CNN) to predict the parameters $w$ of our model for a given image window $I$, jointly with an object classifier. We call this CNN the Normalizer.

The classifier is also a CNN, as deep architectures perform well for this problem, but also to make joint training of the Normalizer and the classifier easy. We refer to this classifier as the Detector. Joint training is done by minimizing the following loss function:

$$(\hat{\Theta}, \hat{\Phi}) = \arg\min_{\Theta, \Phi} \sum_j \ell\left( g^{(\Theta)}\left( \text{ALCN}(I_j;\, f^{(\Phi)}(I_j)) \right); y_j \right), \tag{2}$$

where $\Theta$ and $\Phi$ are the parameters of the Detector CNN $g(\cdot)$ and the Normalizer $f(\cdot)$, respectively; $\ell(\cdot; y)$ is the negative log-likelihood loss function. $I_j$ and $y_j$ are training image regions and their labels: We use image regions extracted from the Phos dataset [24], including the images shown in Fig. 3, the labels are either background or the index of the object contained in the corresponding image region. We use Phos for our purpose because it is made of various objects under different illumination conditions, with 9 images captured under various strengths of uniform illumination and 6 images under non-uniform illumination from various directions. In practice, we use Theano [1] to optimize Eq. (2).

## 3.3 From Window to Image Normalization

Once trained on windows, we apply the Normalizer to the whole input images by extending Eq. (1) to

$$\text{ALCN}(\mathbf{I}) = \sum_{k=1}^{N} G_{\sigma_k^{\text{ALCN}}} * (F_k(\mathbf{I}) \circ \mathbf{I}), \tag{3}$$

where $F_k(\mathbf{I})$ is a weight matrix with the same dimension as the input image $\mathbf{I}$ for the $k$-th 2D Gaussian filter, and $\circ$ is the Hadamard (element-wise) product. The weight matrix $F_k(\mathbf{I})$ corresponding to the $k$-th 2D Gaussian filter is computed as $(F_k(\mathbf{I}))_{ij} = f_k(I_{ij})$, where $(\cdot)_{ij}$ is the entry in the $i$-th row and $j$-th column of the matrix, $I_{ij}$ is the image window centered at $(i, j)$ in image $\mathbf{I}$, and $f_k(\cdot)$ is the $k$-th weight predicted by the Normalizer for the given image window. This can be done very efficiently by sharing the convolutions between windows [6].

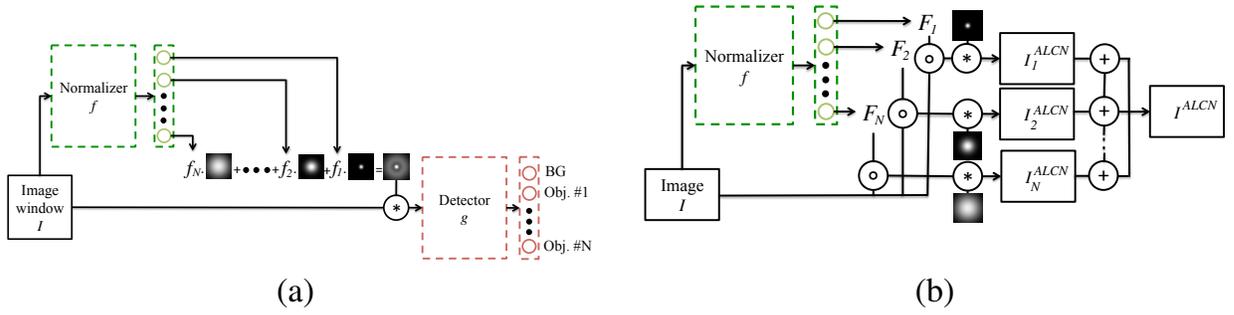

(a)            (b)

Figure 2: Overview of our method. (a) We first train our Normalizer jointly with the Detector using image regions from the Phos dataset. (b) We can then normalize images of previously unseen objects by applying this Normalizer to predict the parameters of our normalization model.

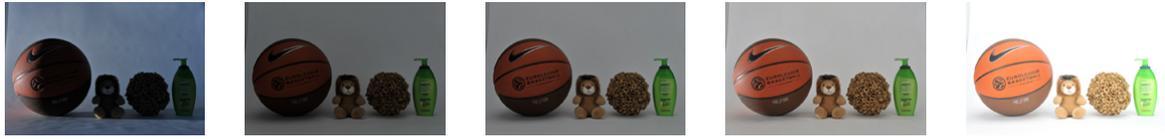

Figure 3: Four of the ten objects we use from the Phos dataset [24] under different illuminations.

Normalization is therefore different for each location of the input image. This allows us to adapt better to the local illumination conditions. Because it relies on Gaussian filtering, it is also fast taking only 50 ms for 10 2D Gaussian filters, on an Intel Core i7-5820K 3.30 GHz desktop with a GeForce GTX 980 Ti on a $128 \times 128$ image.

## 4 Experiments

This section presents the evaluation of our method. We first evaluate it on 2D object detection under drastic illumination changes when only few images are available for training. We then evaluate it for the 2D detection of object parts, when aiming at estimation the object's 3D pose when training images are all captured under the same illumination. In addition, we show that our Normalizer trained for object detection makes 3D pose estimation robust to illumination changes as well.

### 4.1 Datasets

Some datasets have instances captured under different illuminations, such as NORB [13], ALOI [5], CMU Multi-PIE [7] or Phos [24]. However, they are not suitable for our purposes: NORB has only 6 different lighting directions; the images of ALOI contain a single object only and over a black background; CMU Multi-PIE was developed for face recognition and the image is always centered on the face; Phos was useful for our joint training approach, however, it has only 15 test images and the objects are always at the same locations, which would make the evaluation dubious.

Therefore, we created a new dataset for benchmarking object detection under challenging lighting conditions and cluttered background. We will refer to this dataset as the ALCN-2D dataset. As shown in Fig. 4, we selected three objects spanning different material properties: plastic, velvet and metal (velvet has a BRDF that is neither Lambertian nor specular [16], and the metallic object—the watch—is very specular). For each object, we have 10 $300 \times 300$ grayscale training images and 1200 $1280 \times 800$ grayscale test images, exhibiting these objects under different illuminations, different lighting colors, and distractors in the background. The number of test images is therefore much larger than for previous datasets. We manually annotated the ground truth bounding boxes in the test images in which the target object is present. In this first dataset, the objects are intentionally moved on a planar surface, in order to limit the perspective appearance changes and focus on the illumination variations.

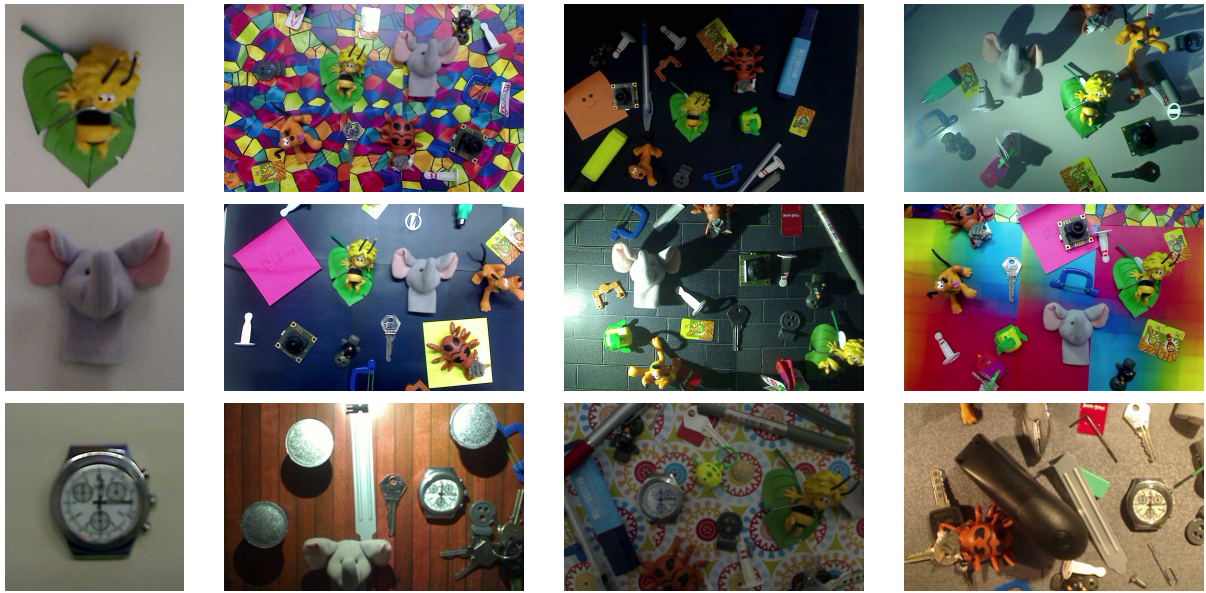

Figure 4: The objects for our ALCN-2D dataset, and representative test images. We selected three objects spanning different material properties: plastic, velvet, metal (velvet has a BRDF that is neither Lambertian nor specular, and the metallic object—the watch—is very specular). By contrast with previous datasets, we have a very large number of test images (1200 for each object), capturing many different illuminations and background.

The second dataset we consider is the Box Dataset from the authors of [3], which combines perspective and light changes. It is made of a registered training sequence of an electric box under various 3D poses but a single illumination and a test sequence of the same box under various 3D poses and illuminations. Some images are shown in the second row of Fig. 1. This test sequence was actually not part of the experiments of [3] as it was too challenging. The goal is to estimate the 3D pose of the box.

Finally, we introduce another dataset for 3D pose estimation. This dataset is made of a training sequence of 1000 registered frames of the Duck from the Hinterstoisser dataset [9] obtained by 3D printing under a single illumination and 8 testing sequences under various illuminations. Some images are shown in the third row of Fig. 1. We will refer to this dataset as the ALCN-Duck dataset.

## 4.2   Network Architecture and Optimization Details

A rather simple architecture is sufficient for the Normalizer: For all our experiments, the first layer performs 20 convolutions with 5×5 filters with 2×2 max-pooling. The second layer performs 50 5×5 convolutions followed by 2×2 max-pooling. The third layer is a fully connected layer of 1024 hidden units. The last layer returns the predicted weights. To keep optimization tractable, we downscaled the training images of the target objects by a factor of 10. To avoid border effects, we use $48 \times 48$ input patches for the Normalizer, and use $32 \times 32$ patches as input to the Detector. We use the tanh function as activation function, as it performs better than ReLU on our problem. This difference is because a sigmoid can control better the large range of intensities exhibited in the images of our dataset, while other datasets have much more controlled illuminations.

Our goal is to reduce the number of required training images in order to be robust to illumination changes. To train the Normalizer on the Phos dataset, we generate synthetic images from real images to recognize these objects. We use a very simple method to generate these images: We add Gaussian noise, replace the background by a random one, which can be uniform or cropped from an image from the ImageNet dataset [4], and apply small affine transformations to the intensities. We use images from the ImageNet dataset [4] as negative samples. We generate 500,000 synthetic images, with the same number of false and negative images. Once the Normalizer is trained on the Phos dataset, we can use synthetic images created from a very small number of real images of the target objects to train a new classifier to recognize these objects: Some of our experiments presented below use only one real image. At test time, we run the Detector on all $48 \times 48$ image windows extracted from the test image.

## 4.3 Experiments and Discussion

For evaluation, we use the PASCAL criterion to decide if a detection is correct with an Intersection over Union of 0.8, with fixed box sizes of $300 \times 300$, reporting Precision-Recall (PR) curves and Areas Under Curve (AUC) in order to compare the performances of the different methods.

### 4.3.1 Explicit Normalization vs Illumination Robustness with Deep Learning

As mentioned in the introduction, Deep Networks can learn robustness to illumination variations without explicitly handling them, at least to some extent. To show that our method allows us to go further, we first tried to train from scratch several Deep Network architectures, without normalizing the images beforehand, by varying the number of layers and the number of filters for each layer. We use one real example of each object in the ALCN-2D dataset for this experiment. The best architecture we found performs with an AUC of 0.606. However, our method still performs better with an AUC of 0.787. This shows that our approach achieves better robustness to illumination than a single CNN, at least when the training set is limited, as in our scenario.

We also evaluated Deep Residual Learning Network architectures [8]. We used the same network architectures and training parameters as in [8] on CIFAR-10. ResNets with 20, 32, 44, 56 and 110 layers perform with AUCs of 0.456, 0.498, 0.518, 0.589 and 0.565 respectively, which is still outperformed by a much simpler network when our normalization is used. Between 56 and 110 layers, the network starts overfitting, and increasing the number of layers results in a decrease of performance.

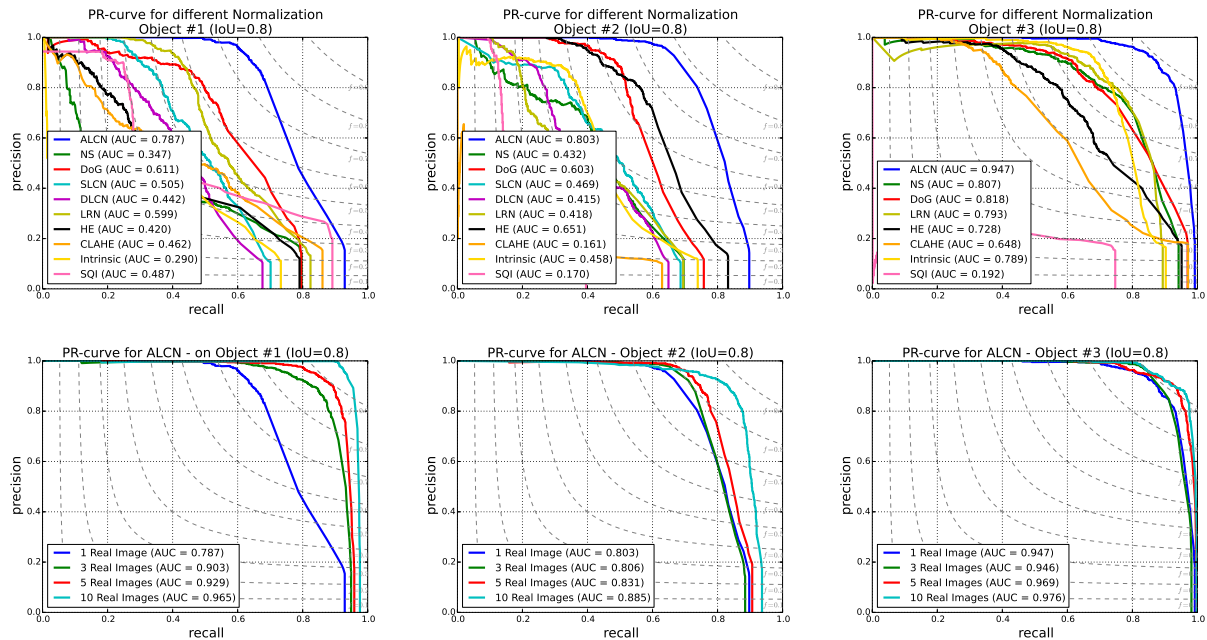

**Figure 5:** **First row**: Comparing different normalization methods using the best parameter values for each method for Objects #1, #2 and #3 of ALCN-2D. ALCN systematically performs best by a large margin. **Second row**: Evaluating the influence of the number of real images used for training the detector on Objects #1, #2 and #3 of ALCN-2D. The detection accuracy keeps increasing when using more real images for generating the training set.

### 4.3.2 Comparing ALCN against previous Normalization Methods

We consider different existing methods in our evaluations: Normalization by Standardization (NS), Difference-of-Gaussians (DoG), Whitening, Subtractive and Divisive Local Contrast Normalization (SLCN and DLCN) [10], Local Response Normalization (LRN) [11], Histogram Equalization (HE), Contrast Limited Adaptive Histogram Equalization (CLAHE) [18], using the Intrinsic Image [20] and Self Quotient Image (SQI) [25].

In order to assess the effects of different normalization techniques on the detection performances, we employed the same detector architecture for the normalization methods, but re-training it for every normalization method. Fig. 5(a) compares these methods on the ALCN-2D dataset. For DoG, Subtractive and Divisive LCN, we optimized their parameters to perform best on the training set. Our method consistently outperforms the others for all objects of the ALCN dataset. Most of the other methods have very different performances across the different objects of the dataset. Whitening obtained an extremely bad score on all datasets, while both versions of LCN failed in detecting Object #3, the most specular object, obtaining an AUC score smaller than 0.1.

### 4.3.3 Number of Training Images

Fig. 5(b) shows how our detector performs on the ALCN-2D dataset for different numbers of real images used for training. To build the full training set, we generate synthetic images from each of these real images, but we keep the total number of synthetic training images constant. Using only 1 image to generate the whole training set already gives us good results. Using more real images improves the performances further, 10 images are enough for very good performance. This shows that we can learn to detect objects under very different drastic

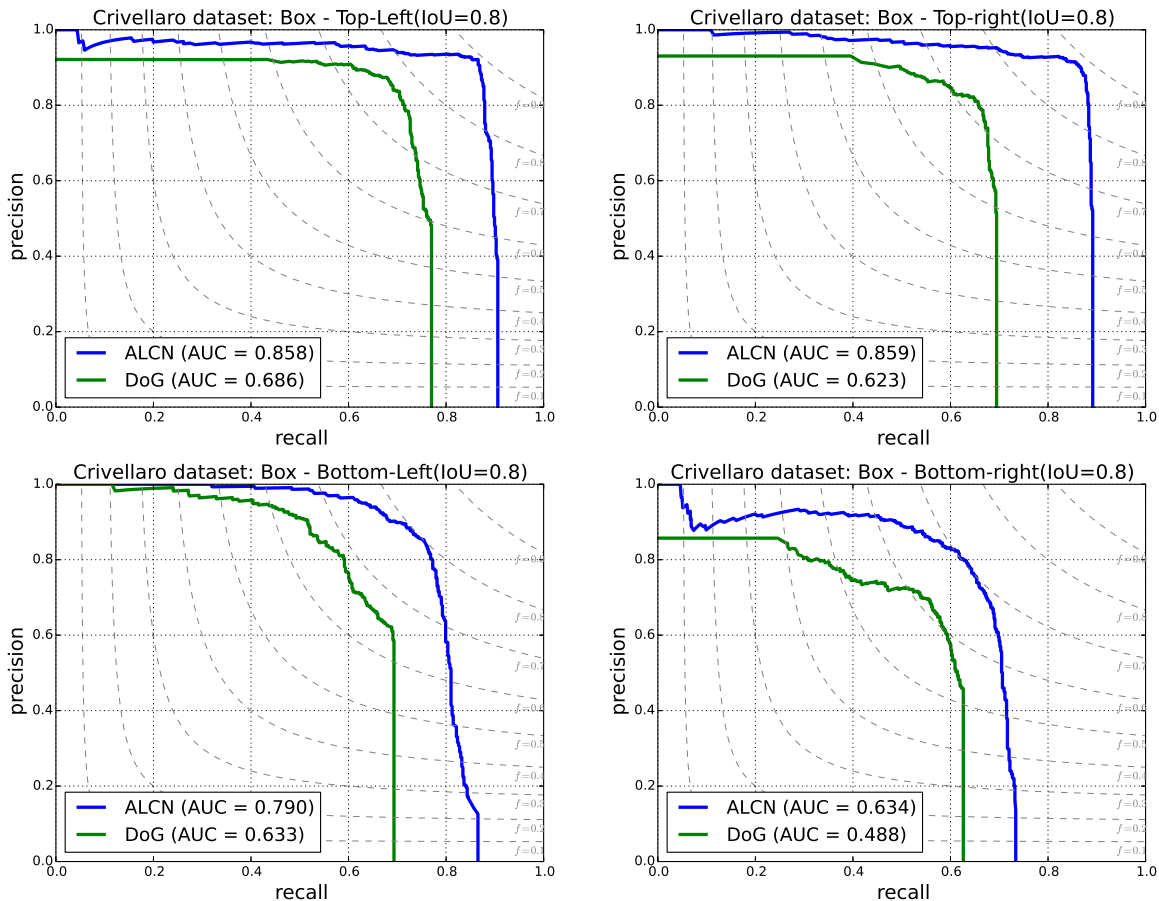

Figure 6: Comparing ALCN and DoG on the BOX dataset - Video #3 from Crivellaro [3]. Our ALCN performs best at detecting the corners of the box.

| sequence | w/o illumination changes #1 | with illumination changes | | | | | | |
|---|---|---|---|---|---|---|---|---|
| | | #2 | #3 | #4 | #5 | #6 | #7 | #8 |
| VGG | **100** | 47.26 | 18.33 | 32.65 | 0.00 | 0.00 | 0.00 | 0.00 |
| VGG+ALCN | **100** | **77.78** | **60.71** | **70.68** | **64.08** | **51.37** | **76.20** | **50.10** |

Table 1: Percentage of correctly estimated poses using the 2D Projection metric of [2], when the method of [19] is applied to our ALCN-Duck sequences, with and without ALCN. Using VGG—trained to predict the 3D pose—alone is not sufficient when illumination changes, ALCN allows us to retrieve accurate poses.

illuminations from very few real examples augmented with simple synthetic examples.

## 4.4 3D Object Pose Estimation

As mentioned in the introduction, our main goal is to train Deep Networks methods for 3D pose estimation, without requiring large amounts of training data while being robust to light changes. We evaluate here ALCN for this goal on two different datasets.

**Box dataset.** To evaluate ALCN for 3D object detection and pose estimation, we first applied it on the Box dataset described in Section 4.1 using the method of [3], which is based on part detection: It first learns to detect some parts of the target object, then it predicts the 3D pose of each part to finally combine them to estimate the object 3D pose.

The test videos from [3] exhibit challenging dynamic complex background and light changes. We changed the code provided by the authors to apply ALCN before the part detection. We evaluated DoG normalization, the second best method according to our previous experiments, optimized on these training images, against our Normalizer. Fig. 6 shows the results; ALCN allows us to detect the parts more robustly and thus to compute much more stable poses.

**ALCN-Duck dataset.** The method proposed in [19] first detects the target object using a detector and then, given the image window centered on the object, predicts the 3D pose of the object using a regressor. For both, detector and regressor, [19] finetunes convolution and fully connected layers of VGG [22], and achieved very good results on the Hinterstoisser dataset. However, this dataset does not exhibit strong light changes, and we evaluated our approach on the ALCN-Duck dataset described in Section 4.1. Here, we use color images as input to the detector and regressor. To apply ALCN to these images, we use a method inspired by [26]: We first transform the input color image in the CIE Lab colorspace, normalize the lightness map L with our method and re-transform the image in the RGB space without changing the ab channels.

Table 1 gives the percentage of correctly estimated poses using the 2D Projection metric [2] with and without our ALCN normalization. [19], with and without ALCN, performs very well on video sequence #1, which has no illumination changes. It performs much worse when ALCN is not used on Sequences #2, #3 and #4, where the illuminations are slightly different from training. For the other sequences, which have much more challenging lightening conditions, it dramatically fails to recover the object poses. This shows that ALCN can provide illumination invariance at a level to which deep networks such as VGG cannot. Some qualitative results are shown on the last row of Fig. 1.

# 5   Conclusion

We proposed an efficient approach to illumination normalization that improve robustness to light changes for object detection and 3D pose estimation methods without requiring many training images. This approach should help bringing the power of Deep Learning to applications for which large amounts of training data are not available.

# Acknowledgment

This work was funded by the Christian Doppler Laboratory for Semantic 3D Computer Vision.